%% file: CameraReady2027.tex
\documentclass[letterpaper]{article} % DO NOT CHANGE THIS
\usepackage{aaai2027}  % DO NOT CHANGE THIS
\usepackage[hyphens]{url}  % DO NOT CHANGE THIS
\usepackage{graphicx} % DO NOT CHANGE THIS
\usepackage{natbib}  % DO NOT CHANGE THIS AND DO NOT ADD ANY OPTIONS TO IT
\usepackage{caption} % DO NOT CHANGE THIS AND DO NOT ADD ANY OPTIONS TO IT
\usepackage{algorithm}
\usepackage{algorithmic}
\usepackage{amsmath}
\usepackage{multirow}
\usepackage{amssymb}
\usepackage{newfloat}
\usepackage{listings}
\DeclareCaptionStyle{ruled}{labelfont=normalfont,labelsep=colon,strut=off} % DO NOT CHANGE THIS
\floatstyle{ruled}
\newfloat{listing}{tb}{lst}{}
\floatname{listing}{Listing}

\usepackage{booktabs}

\title{TwinIR: Coordinated Invisible Dual-Point Attacks on Online HD Map Construction}
\author{
    Haibo Hu$^{1}$\quad Jianghuai Deng\textsuperscript{\rm 1}\thanks{Equal contribution}\quad Chen Tang$^{1}$\quad  Yang Lou\textsuperscript{\rm 1}\thanks{Corresponding author: yanglou-c@my.cityu.edu.hk}\quad  \\Qian Xu$^{1}$\quad  Jianping Wang$^{1}$\quad \\
    $^{1}$City University of Hong Kong
}
\affiliations{
}

\begin{document}
\nocopyright
\maketitle

\begin{abstract}
Online HD map construction is critical to prediction and planning in autonomous driving. We find that existing physical attacks against online map construction are limited by a cross-boundary compensation effect: after the target boundary is perturbed, another visible boundary may retain sufficient geometric cues for the model to recover the original road geometry. Based on this observation, we propose TwinIR, a new mechanism-guided physical attack methodology for online map construction. TwinIR jointly optimizes attack effectiveness and point sparsity, seeking the minimum number of attack points needed to suppress compensating geometric cues from surrounding boundaries. To reduce the perceptibility of multi-point attacks, TwinIR models camera responses to near-infrared illumination and maps optimized attack points to feasible physical placements, producing camera-visible interference with minimal visible-spectrum changes. Experiments on nuScenes across state-of-the-art online map construction models show that TwinIR reduces mAP by 8.18–8.96 percentage points under RSA and 2.84–5.62 points under ETA, while increasing the unreachable-goal rate by 25–28 points and the unsafe-planned-trajectory rate by 19–20 points over clean inputs. These attacks are also validated on a real-world testbed AV, where TwinIR successfully induces both road straightening and early-turn deformations while remaining inconspicuous in full-color views.
% Online HD map construction is critical to prediction and planning in autonomous driving. We find that existing single-point physical attacks can be limited by a \textbf{cross-boundary compensation effect}: after the target boundary is perturbed, another visible boundary may retain sufficient geometric cues for the model to recover the original road geometry. Based on this observation, we propose \textbf{TwinIR}, a new mechanism-guided physical attack methodology for online map construction. TwinIR uses an objective-conditioned search algorithm to perturb the target boundary and introduces a second attack position on the compensation boundary only when it strengthens the intended deformation, while preserving the guiding boundary. It further models camera responses to near-infrared illumination during digital optimization and maps the optimized attack positions to feasible physical deployment positions, producing camera-visible interference with limited visible-spectrum changes. Experiments on nuScenes across state-of-the-art online map construction models show that TwinIR reduces mAP by 8.18--8.96 percentage points under RSA and 2.84--5.62 points under ETA, while increasing the unreachable-goal rate by 25--28 points and the unsafe-planned-trajectory rate by 19--20 points over clean inputs. These attacks are also validated on a real-world testbed AV, where TwinIR successfully induces both road straightening and early-turn deformations while remaining inconspicuous in full-color views.

\end{abstract}
\input{sections/introduction}
\input{sections/motivating_study}

\input{sections/threat_model}

\input{sections/method}

\input{sections/experiments}

\input{sections/conclusion}
\bibliography{aaai2027}

\clearpage
\appendix
\input{sections/related_work}
\input{sections/hardware_appendix}

\end{document}

%% file: sections/introduction.tex
\section{Introduction}

% \begin{figure}[t]
%   \centering
%   \includegraphics[width=\linewidth]{Figures/attack_teaser.pdf}
%   \caption{\textbf{TwinIR attack scenario and targeted effects.}
%   Roadside near-infrared sources create camera-visible interference with
%   limited visible change to human observers. Solid green curves mark actual road
%   boundaries, dashed red curves the attacked predictions, and blue curves
%   the resulting planned trajectories. RSA can omit a valid turn, whereas
%   ETA can induce a premature turn and create boundary-crossing risk.}
%   \label{fig:attack_teaser}
% \end{figure}

High-definition (HD) maps encode lane boundaries, dividers, pedestrian crossings, and other static road structures that support autonomous-driving prediction and planning \cite{li2022hdmapnet,liao2025maptrv2}. Unlike traditional offline construction, which requires large-scale data collection, manual annotation, and continuous maintenance, online HD map construction predicts local vectorized maps directly from onboard observations \cite{li2022hdmapnet,liu2023vectormapnet}. Recent models improve map consistency by aggregating structured point sets, geometric relations, and temporal context \cite{liao2023maptr,liao2025maptrv2,yuan2024streammapnet,chen2024maptracker}.Beyond point-set representations, recent online mapping methods have explored piecewise curve and pivotal-point parameterizations to improve compactness and geometric precision~\cite{qiao2023end,ding2023pivotnet}. Because their predictions directly guide motion planning, errors induced by physical interference can propagate to safety-critical driving decisions~\cite{hu2022st,hu2023planning,jiang2023vad}.
\begin{figure}[t]
\centering
\includegraphics[width=\linewidth, trim=0cm 9cm 17cm 0cm, clip]{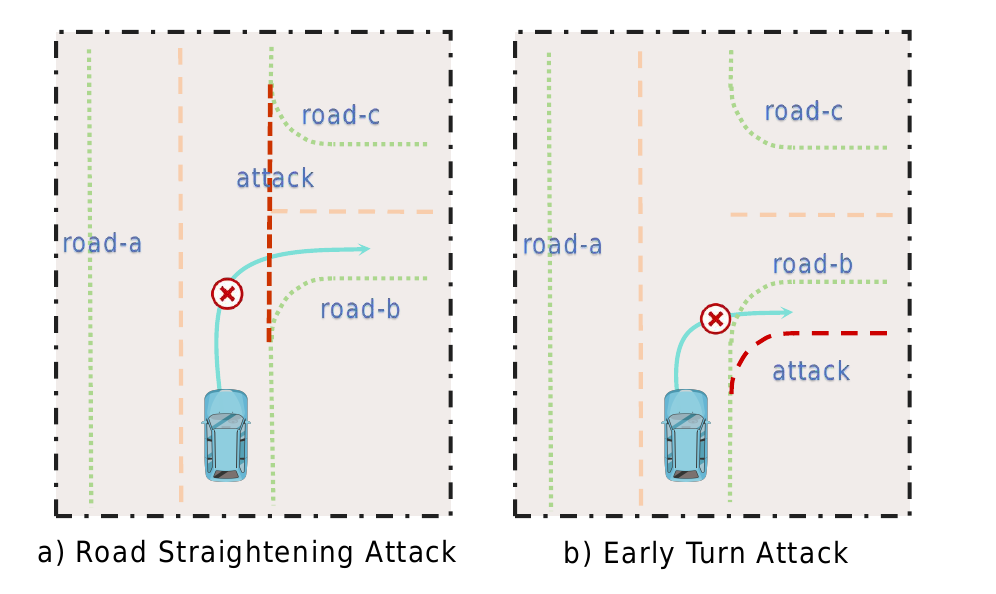}
\caption{\textbf{Objective-specific boundary roles.}
In both panels, \textit{road-b} is the target boundary.}
\label{fig:single_fail_case}
\end{figure}

Despite the rapid progress of online HD map construction, its security under physical-world interference remains insufficiently understood. Physical adversarial attacks have been studied across image classification, object detection, LiDAR perception, and BEV-based autonomous-driving systems, with recent work emphasizing that digital attack effectiveness does not necessarily translate into physical or driving-level impact~\cite{eykholt2018robust,tu2020physically,zhu2023understanding,hingun2023reap,wang2023does}. More recently, the first systematic security analysis of online HD map construction reveals a bias toward symmetric predictions in asymmetric road scenes~\cite{lou2025asymmetry}. At a fork, disrupting geometric cues near the turn can cause a diverging boundary to be predicted as straight, thereby converting an asymmetric fork into a symmetric road structure. Based on this bias, the work introduces a single-position blinding attack and an adversarial-patch attack, including the Road Straightening Attack (RSA), which removes a predicted turn, and the Early Turn Attack (ETA), which shifts the predicted turn toward the ego vehicle. As illustrated in Fig.~\ref{fig:single_fail_case}, these map errors can further propagate to motion planning. However, in some scenes, the single-point design leaves complementary boundary cues intact, allowing the model to preserve or recover the original road geometry.

Our analysis of unsuccessful cases explains why searching additional candidate positions does not always improve a single-point attack. We call the boundary whose prediction the attacker seeks to change the \textbf{target boundary}. Prior work uses \emph{reference boundary} for a non-target boundary that provides contextual cues to this prediction. We refine this relation by function: a \textbf{guiding boundary} supports the desired deformation, whereas a \textbf{compensation boundary} preserves the original geometry (Fig.~\ref{fig:single_fail_case}). If an attack position perturbs the target boundary but leaves the compensation boundary visible, the model can recover the original turn; we call this behavior the \textbf{cross-boundary compensation effect}. This observation indicates that a reliable attack may require coordinated interference at both the target and compensation boundaries, rather than repeatedly searching for a stronger single-point perturbation.

However, such multi-point attacks introduce a practical challenge: deploying multiple visible flashlights or adversarial patches increases the number of physical sources and makes the attack substantially more conspicuous. To preserve the effectiveness of coordinated interference while reducing visual exposure, we explore near-infrared illumination, which is weakly perceived by humans but remains detectable by camera sensors~\cite{wang2021icansee,sato2024invisible}. We propose \textbf{TwinIR}, a mechanism-guided, at-most-two-point attack framework. Its \textbf{objective-conditioned attack-position localization} first identifies an attack position on the target boundary and retains a second position on the compensation boundary only when it provides additional attack benefit, while keeping the guiding boundary visible. Its \textbf{near-infrared realization} simulates the camera response during digital search and transfers the retained attack positions to feasible roadside deployment locations using ego-relative coordinates.

We evaluate TwinIR on the nuScenes asymmetric-scene subset against MapTR, MGMap, and DAMap under the same query budget as the single-point attack. Compared with clean inputs, TwinIR reduces mAP by 8.18--8.96 percentage
points under RSA and by 2.84--5.62 points under ETA. It also increases the unreachable-goal rate by 11--12 points and the unsafe planned trajectory rate by 3--8 points over the single-point attack. In a controlled physical comparison, the near-infrared source produces a clear response in an infrared-sensitive camera with little visible change in full-color views.

Our main contributions are summarized as follows:

\begin{itemize}
\item We identify the \textbf{cross-boundary compensation effect}: existing attacks on online map construction can fail because an unaffected compensation boundary preserves the original road geometry after a single-point attack.
\item We propose \textbf{TwinIR}, a mechanism-guided attack framework that uses at most two attack positions. TwinIR first localizes an attack position on the target boundary and conditionally retains a second attack position on the objective-specific compensation boundary only when it improves the attack, while keeping the guiding boundary visible. The selected positions are realized using near-infrared illumination to minimize visible-spectrum changes.
\item We evaluate TwinIR against state-of-the-art online map construction models on the nuScenes dataset and in real-world physical experiments. TwinIR consistently causes greater map deformation and downstream planning disruption than single-point blinding while remaining visually covert.
\end{itemize}

%% file: sections/motivating_study.tex
\section{Motivation Study: Cross-Boundary Compensation}
\label{sec:method_motivation}

% Prior work identifies a symmetry bias in online HD map construction~\cite{lou2025asymmetry}. At an asymmetric fork, one road boundary diverges while the opposite boundary remains relatively straight. When geometric cues near the turn are disrupted, the model may incorrectly straighten the diverging boundary, transforming the fork into a symmetric road. Although a single optimized blinding position or adversarial patch can trigger this behavior, its effectiveness remains limited: at most 44\% of targeted routes become unreachable, and the unsafe planned trajectory rate reaches at most 27\%. These failures suggest that perturbing a single attack position does not consistently remove all geometric cues supporting the original road structure.
In some asymmetric road scenes, a single attack point cannot reliably induce the intended map deformation. Although the local turning region is perturbed, other visible boundaries may still preserve sufficient geometric cues for the model to reconstruct the original road structure. This suggests that single-point attacks do not consistently suppress all evidence supporting the target prediction.

\subsection{Why Single-Point Attacks Fail}

\begin{figure}[t]
\centering
\includegraphics[trim={0cm 5cm 2.3cm 0cm}, clip, width=\linewidth]{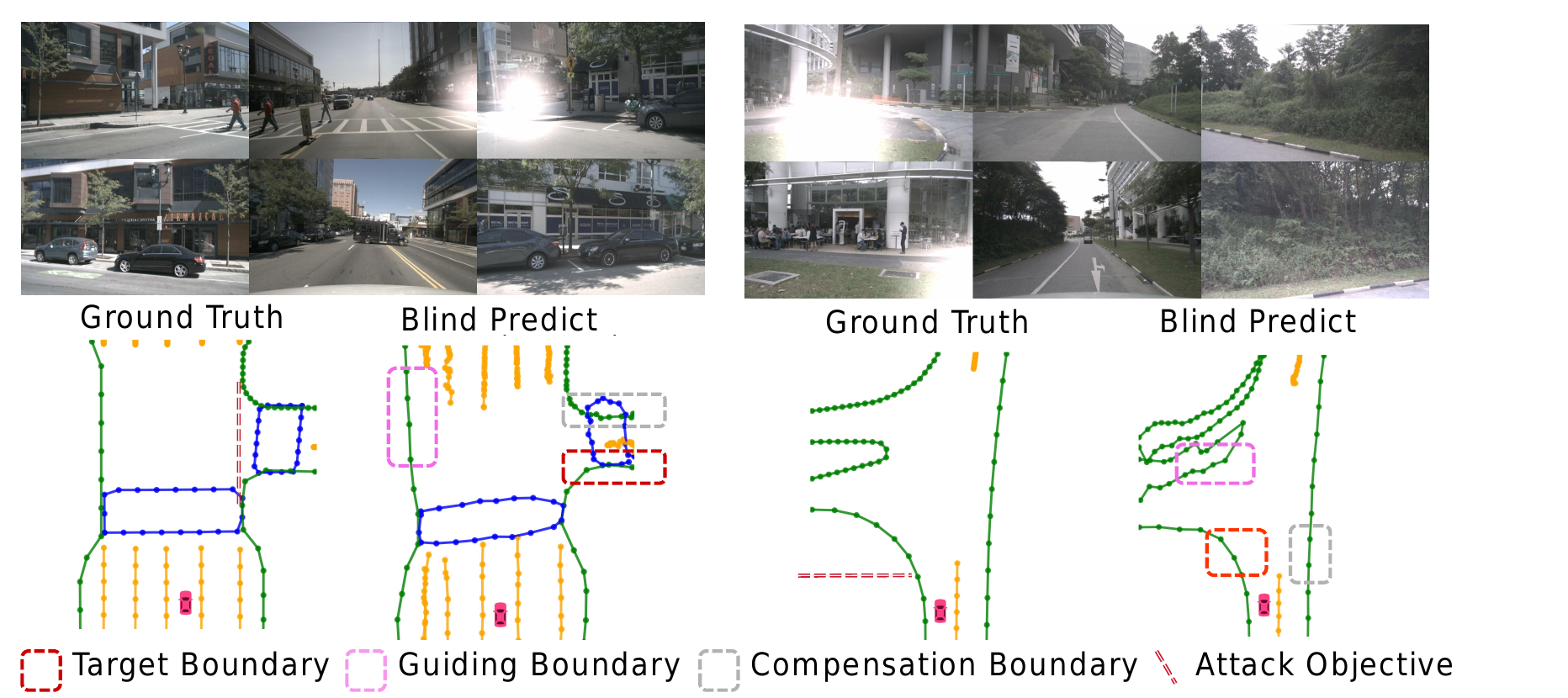}
\caption{\textbf{A representative failure of a single-point attack.}
Although the perturbation affects one side of the turning road, the model preserves the original turn using geometric cues from another visible boundary.}
\label{fig:motivation}
\end{figure}

Figure~\ref{fig:motivation} shows a representative failure. The blinding position perturbs one side of the turning road, yet the predicted map still preserves the original turn.

Across this case and other failures, we observe a recurring pattern: the perturbation removes local cues from the attacked boundary, but a spatially separated boundary remains visible and preserves the curvature or turning-location cues needed to reconstruct the original geometry. Because a single attack position cannot cover both regions, searching additional candidate positions under the same one-point constraint cannot eliminate these residual cues.

We refer to the road boundary whose prediction the attacker seeks to alter as the \textbf{target boundary}. Prior work uses the term \emph{reference boundary} for a non-target boundary that provides contextual cues for predicting the target~\cite{lou2025asymmetry}. We further distinguish two roles among such boundaries:

\begin{itemize}
\item A \textbf{guiding boundary} provides geometric cues consistent with the intended deformation and should remain visible.
\item A \textbf{compensation boundary} provides geometric cues from which the model can reconstruct the original geometry and is therefore a candidate for additional perturbation.
\end{itemize}

In Fig.~\ref{fig:motivation}, although the target boundary is perturbed, a visible compensation boundary enables the model to recover the original turn. We call this behavior the \textbf{cross-boundary compensation effect}. It explains why optimizing a single attack position is insufficient when the cues preserving the original prediction lie on another boundary.
To examine whether the single-point limitation has a measurable downstream
impact, we compare the planning results of single- and dual-point attacks
averaged across three victim models. As shown in
Table~\ref{tab:motivation_planning}, the dual-point variant increases UGR
from 41.67\% to 53.00\% for RSA and UPTR from 26.00\% to 31.67\% for ETA.
These consistent gains indicate that suppressing an additional reference
boundary enables more reliable targeted map deformation.
% To determine whether this effect extends beyond a single case, we compare single- and dual-point attacks on the same 100 frames in the nuScenes dataset~\cite{caesar2020nuscenes} across three models under an identical query budget. As shown in Table~\ref{tab:motivation_stats}, the dual-point variant increases the mAP reduction by 1.21--2.24 percentage points for RSA and 1.78--2.35 points for ETA. The consistent gains across all models and both objectives indicate that residual cross-boundary cues can limit single-point attacks in different online map construction models.

% \begin{table}[t]
% \centering
% \small
% \begin{tabular}{l|cc|cc}
% \toprule
% \multirow{2}{*}{\textbf{Model}} &
% \multicolumn{2}{c|}{\textbf{RSA}} &
% \multicolumn{2}{c}{\textbf{ETA}} \\
% \cmidrule(lr){2-3}\cmidrule(lr){4-5}
% & \textbf{Single} & \textbf{Dual} & \textbf{Single} & \textbf{Dual} \\
% \midrule
% MGMap & 6.97 & 8.18 & 0.81 & 2.84 \\
% MapTR & 6.99 & 8.96 & 2.67 & 5.02 \\
% DAMap & 6.51 & 8.75 & 3.85 & 5.62 \\
% \bottomrule
% \end{tabular}
% \caption{\textbf{Motivating comparison of single- and dual-point
% attacks.} Each entry reports the mAP reduction in percentage points
% relative to the clean input on the same asymmetric scenes and under the
% same query budget. A larger value indicates a stronger attack effect.}
% \label{tab:motivation_stats}
% \end{table}
\begin{table}[t]
\centering
\small
\begin{tabular}{l|ccc}
\toprule
\textbf{Objective} &
\textbf{Single Point} &
\textbf{Dual Point} &
\textbf{Gain (pp)} \\
\midrule
RSA (UGR $\uparrow$)  & 41.67 & 53.00 & +11.33 \\
ETA (UPTR $\uparrow$) & 26.00 & 31.67 & +5.67 \\
\bottomrule
\end{tabular}
\caption{
\textbf{Motivating comparison of single- and dual-point attacks.}
}
\label{tab:motivation_planning}
\end{table}
\subsection{From Cross-Boundary Compensation to Sparse Multi-Point Attacks}

The cross-boundary compensation effect shifts the attack objective from finding the strongest single position to identifying a small set of positions that suppresses the cues preserving the original geometry. Since each physical source increases deployment cost and visibility, TwinIR adopts a sparse two-point design: it perturbs the target boundary and adds a second point on the compensation boundary only when this further improves the attack, while preserving the guiding boundary.

Figure~\ref{fig:single_fail_case} shows how these roles differ across objectives.
\begin{itemize}
    \item \textbf{RSA.} In Fig.~\ref{fig:single_fail_case}(a), \textit{road-b} is the target, curved \textit{road-c} is the compensation boundary, and straight \textit{road-a} is the guiding boundary. TwinIR perturbs \textit{road-b} and \textit{road-c} while leaving \textit{road-a} visible to support straightening.
    \item \textbf{ETA.} In Fig.~\ref{fig:single_fail_case}(b), \textit{road-b} is the target, straight \textit{road-a} is the compensation boundary, and post-turn \textit{road-c} is the guiding boundary. TwinIR perturbs \textit{road-b} and \textit{road-a} while preserving \textit{road-c} to encourage an earlier turn.
\end{itemize}

%% file: sections/threat_model.tex
\section{Threat Model}

We adopt the roadside physical scenario. A victim autonomous vehicle approaches an asymmetric road scene, such as a fork or turn, and constructs a local vectorized map from six surround-view cameras. Before the vehicle arrives, the attacker places near-infrared sources at feasible roadside locations with a line of sight to the cameras observing the relevant road boundaries.

\subsection{Problem Formulation and Attack Objectives}
\textbf{Problem Formulation} Given the multi-view observations $\mathcal{I}=\{I^{c}\}_{c=1}^{C}$, where $C=6$ in our setting, the victim online map construction model $M$ predicts a local vectorized map $\hat{\mathcal{M}}=M(\mathcal{I})$. Each map element is represented as an ordered set of bird's-eye-view points. We consider a targeted attack against a selected road boundary rather than indiscriminately degrading the entire map.

\textbf{Attack Goals} The attacker has two ordered objectives: first maximize the targeted map deformation and then minimize the number of attack positions needed to achieve it. For attack objective $o$, let $\mathcal{P}$ be a set of attack positions drawn from the feasible roadside region $\Omega$, and let $K$ be the maximum attack-position budget. Because TwinIR deploys one near-infrared source at each retained attack position, $|\mathcal{P}|$ also equals the number of physical sources. We write $\mathcal{I}_{\mathcal{P}}$ for the camera observations after applying the perturbations at positions in $\mathcal{P}$ and formulate the problem as
\begin{equation}
    \mathcal{P}^{*}
    =
    \operatorname*{arg\,min}^{\mathrm{lex}}_{
        \mathcal{P}\subseteq\Omega,\ |\mathcal{P}|\leq K
    }
    \left[
        \mathcal{L}_{o}\!\left(M(\mathcal{I}_{\mathcal{P}}),T\right),
        |\mathcal{P}|
    \right],
    \label{eq:dual_point_objective}
\end{equation}
where $T$ is the desired boundary geometry and a smaller $\mathcal{L}_{o}$ denotes a stronger attack. The lexicographic order first minimizes the objective-specific attack loss and then prefers fewer attack positions among configurations with the same attack effect. TwinIR instantiates this formulation with $K=2$: it first selects a primary attack position on the target boundary and retains a secondary attack position on the compensation boundary only when the two-point configuration further reduces the loss.
We consider two targeted objectives. Road Straightening Attack (RSA) makes a curved boundary appear straight and can remove a valid turn from the predicted map. Early Turn Attack (ETA) shifts the predicted turn toward the ego vehicle and can move the planned trajectory across the actual road boundary.

\subsection{Attacker Capabilities and Constraints}
We follow the offline black-box setting and separate the attacker's offline access from the constraints on physical deployment.

\textbf{Offline search.} The attacker has no real-time access to the victim autonomous vehicle. Before deployment, the attacker has pre-collected sensor data from the target scene and obtains the corresponding clean map prediction from the victim model. These observations and predictions are used to optimize the attack configuration.

\textbf{Black-box access.} During offline search, the attacker can query the victim online map construction model with digitally perturbed observations and observe its predicted vectorized map. The attacker cannot access the model architecture, parameters, or gradients.
\textbf{Deployment constraints.} Before the vehicle arrives, the attacker can place at most $K$ near-infrared sources within the feasible roadside region $\Omega$; TwinIR uses $K=2$. Each source must maintain a line of sight to the relevant onboard camera. The attacker cannot modify the vehicle, model, sensor firmware, communication pipeline, or road surface and has no real-time control over the vehicle.

%% file: sections/method.tex
\section{Method}
\label{sec:method}
\label{sec:overview}
\begin{figure}[b]
  \centering
  \includegraphics[width=\linewidth]{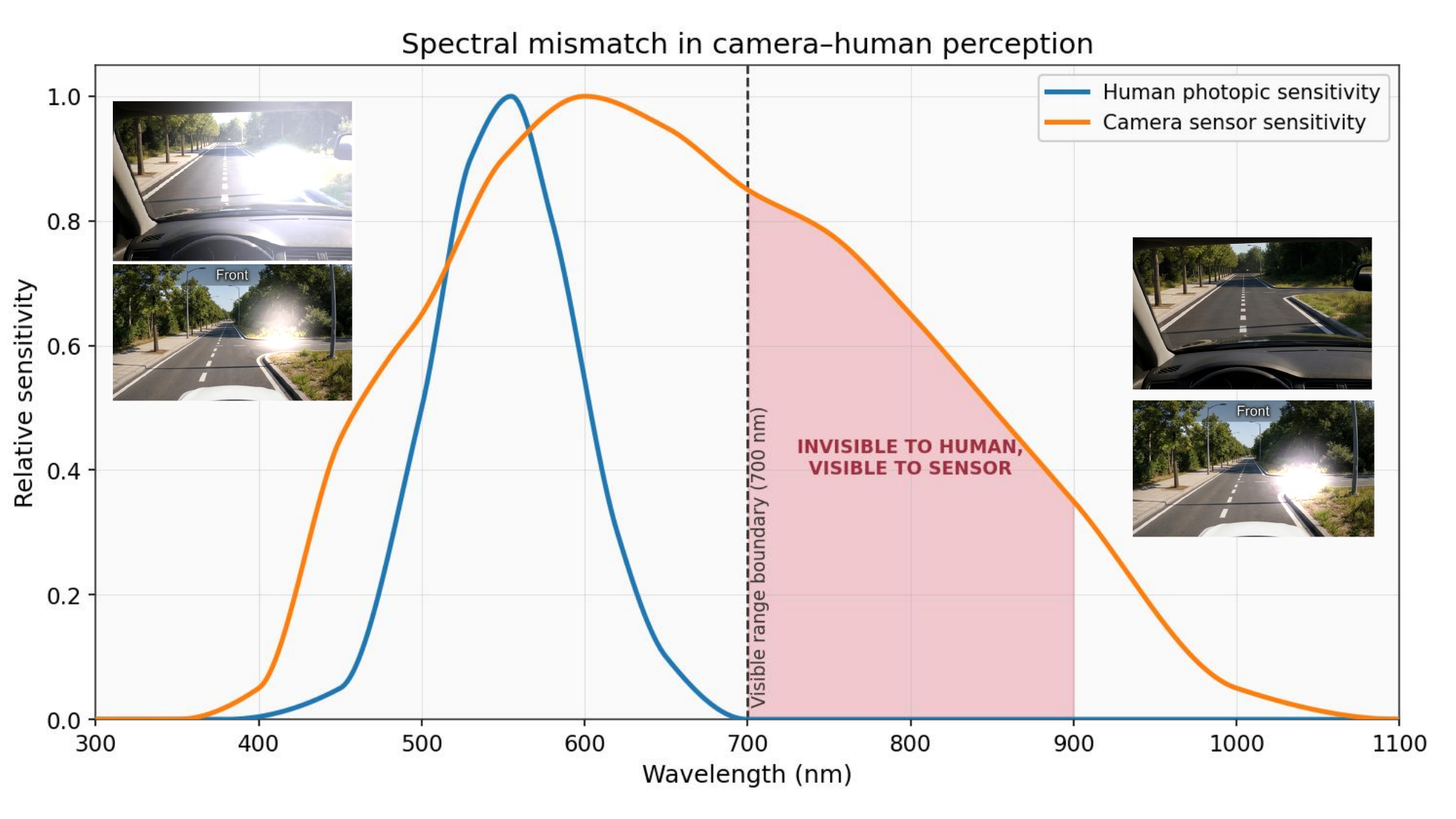}
  \caption{\textbf{Physical basis of spectral stealthiness.} Human
  photopic sensitivity drops sharply beyond the visible spectrum, while
  CMOS sensors remain sensitive to near-infrared light
  ($700$--$1100$nm).}
  \label{fig:spectral_mismatch}
\end{figure}
\begin{figure*}[!t]
\centering
\includegraphics[width=0.8\textwidth, trim=0 4.5cm 4cm 0, clip]{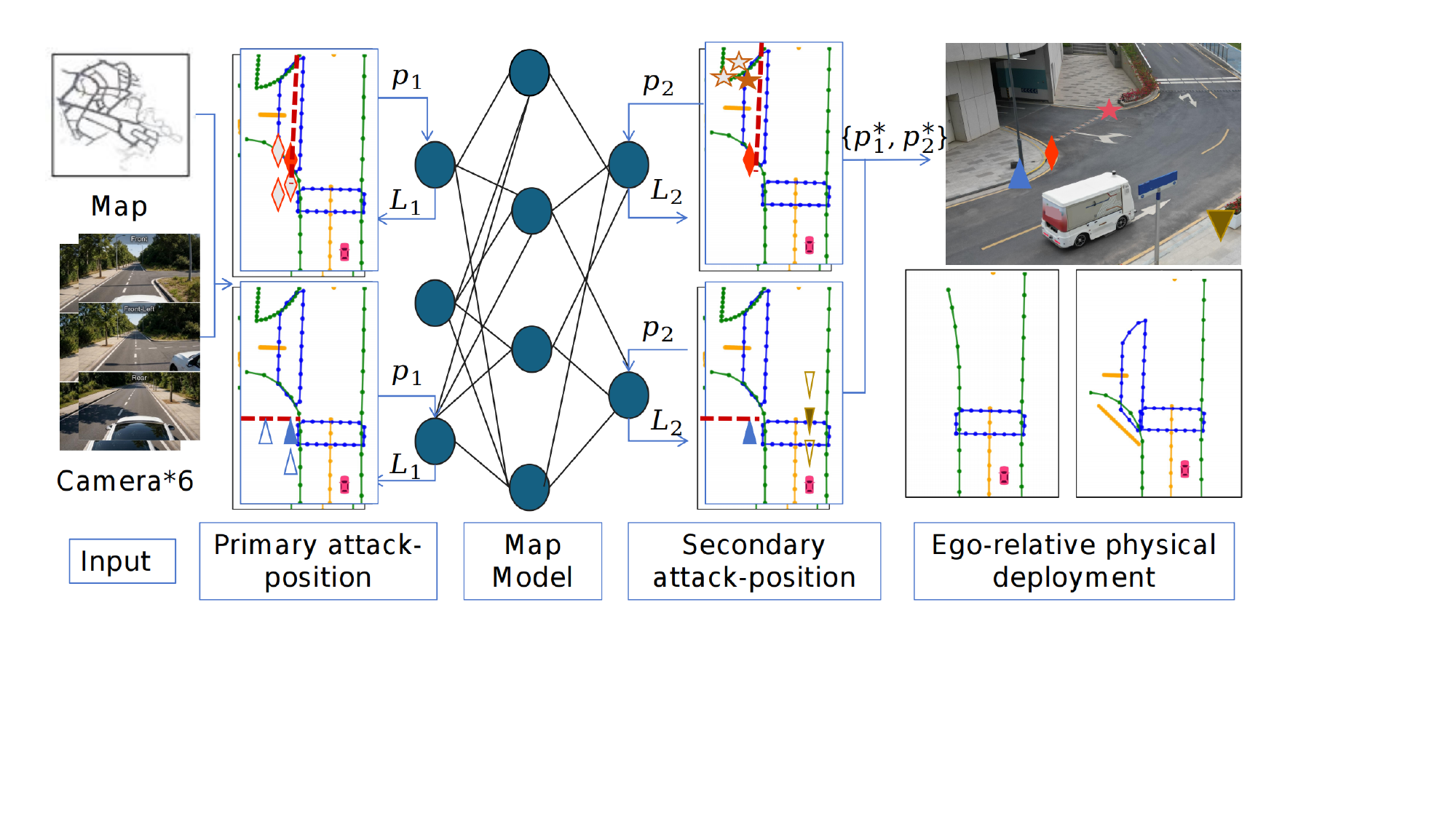}
\vspace{-15pt}
\caption{\textbf{Overview of the TwinIR workflow.}}
\label{fig:overview}
\end{figure*}

TwinIR consists of two components. \textbf{Objective-conditioned attack-position localization} solves the sparse attack-position optimization in Eq.~\ref{eq:dual_point_objective}: it determines attack positions that maximize the targeted map deformation and then minimizes the number of retained attack positions. \textbf{Near-infrared realization} reduces the visibility of multi-point interference by modeling its camera response during digital search and transferring the optimized attack positions to feasible roadside deployment positions. Figure~\ref{fig:overview} summarizes the attack pipeline.

\subsection{Objective-Conditioned Attack-Position Localization}
\label{sec:dual_point_localization}
\label{sec:search}

For each scene, TwinIR first assigns the relevant road boundaries to the target, compensation, and guiding roles. It then localizes a primary attack position on the target boundary and an optional secondary attack position on the compensation boundary.

\paragraph{Boundary-role assignment.}
The attack objective and clean-map topology determine three boundary roles: the boundary to perturb, the boundary whose cues may restore the original geometry, and the boundary whose cues support the desired deformation. We denote them by $B_{\mathrm{tgt}}$, $B_{\mathrm{cmp}}^o$, and $B_{\mathrm{g}}^o$, respectively. Curvature does not identify the target boundary; it only locates geometric segments after this assignment. Straight segments have near-zero discrete curvature, whereas a sustained rise marks the turn onset and the following nonzero-curvature segment forms the turning region. Together with connectivity, this profile distinguishes the straight, curved, and post-turn boundaries used below. For RSA, the opposite-side straight boundary is the guiding boundary, while the paired curved boundary is the compensation boundary. For ETA, the farther post-turn boundary is the guiding boundary, while the opposite-side straight boundary is the compensation boundary.
\begin{algorithm}[t]
\caption{Sequential Objective-Conditioned Attack-Position Selection}
\label{alg:rsa_eta_selection}
\begin{algorithmic}[1]
\REQUIRE Images $\mathcal{I}$, victim model $M$, target boundary
$B_{\mathrm{tgt}}$, compensation boundary $B_{\mathrm{cmp}}^o$,
target specification $T$, objective $o\in\{\mathrm{RSA},\mathrm{ETA}\}$
\ENSURE Selected attack positions $\mathcal{P}^*$

\STATE $\mathcal{C}_1\gets
\mathrm{VisibleSamples}(B_{\mathrm{tgt}})$
\STATE $p_1^*\gets\operatorname*{arg\,min}_{p\in\mathcal{C}_1}
\mathcal{L}_o(M(\mathcal{T}_{\mathrm{IR}}(\mathcal{I};p)),T)$
\STATE $\mathcal{I}_1\gets\mathcal{T}_{\mathrm{IR}}(\mathcal{I};p_1^*)$,
$\mathcal{L}_1\gets\mathcal{L}_o(M(\mathcal{I}_1),T)$

\IF{$o=\mathrm{RSA}$}
    \STATE $\mathcal{C}_2\gets
    \mathrm{RSACompensationCandidates}(B_{\mathrm{cmp}}^o,p_1^*)$
\ELSE
    \STATE $\mathcal{C}_2\gets
    \mathrm{UniformSample}(\mathrm{StraightSegment}(B_{\mathrm{cmp}}^o))$
\ENDIF

\IF{$\mathcal{C}_2=\varnothing$}
    \RETURN $\{p_1^*\}$
\ENDIF

\STATE $p_2^*\gets\operatorname*{arg\,min}_{p\in\mathcal{C}_2}
\mathcal{L}_o(M(\mathcal{T}_{\mathrm{IR}}(\mathcal{I}_1;p)),T)$
\STATE $\mathcal{L}_2\gets
\mathcal{L}_o(M(\mathcal{T}_{\mathrm{IR}}(\mathcal{I}_1;p_2^*)),T)$

\IF{$\mathcal{L}_2<\mathcal{L}_1$}
    \RETURN $\{p_1^*,p_2^*\}$
\ELSE
    \RETURN $\{p_1^*\}$
\ENDIF
\end{algorithmic}
\end{algorithm}
\paragraph{Attack-loss construction.}
After assigning the boundary roles, we construct the target specification and attack loss used to rank both primary and secondary candidates. Let $\widehat{B}$ denote the predicted target boundary under a candidate perturbation.
\begin{itemize}
\item \textbf{RSA.} We construct a straightened target $T_{\mathrm{RSA}}$ that preserves $B_{\mathrm{tgt}}$ up to the turn onset and then follows the straight guiding boundary shifted by the average road width. The loss is the Chamfer distance between the attacked prediction and this target:
\begin{equation}
    \mathcal{L}_{\mathrm{RSA}}
    =
    \mathcal{L}_{\mathrm{chamfer}}
    \left(\widehat{B},T_{\mathrm{RSA}}\right).
    \label{eq:rsa_attack_loss}
\end{equation}
\item \textbf{ETA.} Let $b_i$, $c_i$, and $\widehat{b}_i$, for $i=1,\ldots,N$, be corresponding points on the clean target boundary, its adjacent lane centerline, and the attacked prediction. The outward direction and displacement are $\mathbf{d}_i=(b_i-c_i)/\|b_i-c_i\|_2$ and $\Delta_i=(\widehat{b}_i-b_i)^\top\mathbf{d}_i$. The directional loss rewards displacement toward the roadside and penalizes displacement toward the drivable area:
\begin{equation}
    \mathcal{L}_{\mathrm{ETA}}
    =
    \frac{1}{N}\sum_{i=1}^{N}
    \left[
        -\lambda_{\mathrm{out}}\,\mathrm{ReLU}(\Delta_i)
        +\lambda_{\mathrm{in}}\,\mathrm{ReLU}(-\Delta_i)
    \right].
    \label{eq:eta_attack_loss}
\end{equation}
Here, $\lambda_{\mathrm{out}}$ and $\lambda_{\mathrm{in}}$ balance the outward reward and inward penalty.
\end{itemize}
In the following, $\mathcal{L}_o$ denotes the corresponding loss for objective $o\in\{\mathrm{RSA},\mathrm{ETA}\}$.

We formulate dual-point localization as a sequential black-box search. For
each candidate position, we render the infrared marker, query the victim
model, and evaluate the objective-specific loss
$\mathcal{L}_o$, which measures the distance to the desired straightened
boundary for RSA and the displacement of the turning position for ETA.
The primary point $p_1^*$ is selected by enumerating the top-$K_1$
visibility-ranked candidates along the target boundary. After fixing
$p_1^*$, we construct an objective-conditioned secondary candidate set:
RSA samples nearby positions on the compensation boundary with similar
curvature, whereas ETA samples the straight segment of its reference
boundary. The secondary point $p_2^*$ is then selected by directly
evaluating the joint dual-point attack loss, and is retained only when it
improves over the single-point result.

\paragraph{Primary attack-position localization.}
We sample camera-visible candidates $\mathcal{C}_1$ along the target boundary and query the victim model after rendering each candidate. The primary attack position is the candidate that minimizes the objective-specific loss:
\begin{equation}
    p_1^*
    =
    \operatorname*{arg\,min}_{p\in\mathcal{C}_1}
    \mathcal{L}_o
    \left(
        M\left(\mathcal{T}_{\mathrm{IR}}(\mathcal{I};p)\right)
    \right).
    \label{eq:primary_attack_position}
\end{equation}
We denote the resulting single-position loss by $\mathcal{L}_1$. Thus, the objective-specific loss selects the primary position over the target boundary rather than relying on a manually preferred region.
\begin{table*}[t]
\centering
\resizebox{\textwidth}{!}{
\begin{tabular}{l|cccc|cccc|cccc|cc}
\toprule
\multirow{2}{*}{\textbf{Model}} &
\multicolumn{4}{c|}{\textbf{Clean}} &
\multicolumn{4}{c|}{\textbf{Blind (RSA)}} &
\multicolumn{4}{c|}{\textbf{TwinIR (Ours)}} &
\multicolumn{2}{c}{\textbf{$\Delta$ mAP (pp)}} \\
\cmidrule(lr){2-5}\cmidrule(lr){6-9}\cmidrule(lr){10-13}\cmidrule(lr){14-15}
& Div. & Ped. & Bound. & mAP
& Div. & Ped. & Bound. & mAP
& Div. & Ped. & Bound. & mAP
& Blind-Clean & TwinIR-Clean \\
\midrule
MGMap     & 65.89 & 58.49 & 54.41 & 59.60 & 57.75 & 53.74 & 46.40 & 52.63 & 56.33 & 55.53 & 42.39 & 51.41 & -6.97 & -8.18 \\
MapTR     & 54.18 & 38.17 & 48.95 & 47.10 & 44.12 & 36.06 & 40.16 & 40.11 & 42.77 & 34.15 & 37.51 & 38.14 & -6.99 & -8.96 \\
DAMap     & 69.66 & 59.13 & 53.30 & 60.69 & 61.64 & 55.58 & 45.34 & 54.18 & 60.14 & 53.61 & 42.08 & 51.94 & -6.51 & -8.75 \\
\bottomrule
\end{tabular}
}
\caption{Map element AP(\%) under RSA attacks on asymmetric scenes. Div.=divider, Ped.=ped\_crossing, Bound.=boundary. $\Delta$ mAP is computed in percentage points (pp) w.r.t. Clean.}
\label{tab:rsa_main_full}
\end{table*}
\begin{table}[t]
\centering
\begin{tabular}{l|c|c|c}
\toprule
\textbf{Model} & \textbf{UGR$_c$} & \textbf{Blind (UGR$_c$)} & \textbf{TwinIR (Ours)} \\
\midrule
MGMap     & 26.0 & 40.0 & 51.0 \\
MapTR     & 27.0 & 44.0 & 55.0 \\
DAMap     & 26.0 & 41.0 & 53.0 \\
\bottomrule
\end{tabular}
\caption{Planning impact of RSA on asymmetric scenes, measured by unreachable goal rate (UGR, \%). UGR$_c$ denotes the clean setting.}
\label{tab:rsa_plan}
\end{table}

\paragraph{Secondary attack-position localization.}
With $p_1^*$ fixed, this step searches the compensation boundary for the secondary attack position $p_2^*$ that minimizes the attack loss when combined with $p_1^*$. It locates the best secondary candidate but does not yet decide whether a second position is necessary.
\begin{itemize}
\item \textbf{RSA.} We search the curved compensation boundary for a candidate position that is spatially separated from $p_1^*$ but has curvature similar to the primary region. Separation avoids repeatedly perturbing the same local region, while curvature similarity focuses the search on cues that can restore the original turn.
\item \textbf{ETA.} We search the straight segment of the compensation boundary. Perturbing this segment together with the target boundary's turning region weakens the cues that anchor the original turning position, while the farther post-turn guiding boundary remains visible.
\end{itemize}

\paragraph{Conditional secondary-position selection.}
Secondary attack-position localization produces the strongest two-point candidate, whereas this final step determines whether it justifies an additional physical position. Let $\mathcal{L}_2$ denote the loss obtained by adding $p_2^*$ to the primary perturbation. If $\mathcal{L}_2\geq\mathcal{L}_1$, the secondary attack position is discarded; otherwise, the stronger two-point configuration is retained. This decision implements the lexicographic objective over $k\in\{1,2\}$. Algorithm~\ref{alg:rsa_eta_selection} summarizes the procedure.

\subsection{Near-infrared realization}
Multiple visible-light sources would make the dual-point attack conspicuous.
TwinIR therefore uses near-infrared illumination, which is less perceptible
to humans but remains detectable by vehicle cameras
(Fig.~\ref{fig:spectral_mismatch}). Candidate positions are first evaluated
through digital rendering and then mapped to ego-relative physical locations.

\paragraph{Digital infrared rendering.}
For a candidate point projected to image position $u$, we synthesize a
white core with a Gaussian-blurred purple halo. The resulting infrared
attack spot is
\begin{equation}
    \mathcal{A}(\mathbf{x};u,d)
    =
    \alpha D(\mathbf{x};u,d)\mathbf{c}_{w}
    +
    \beta H(\mathbf{x};u,d)\mathbf{c}_{p},
    \label{eq:ir_marker}
\end{equation}
where $D$ is a distance-dependent white disk, $H$ is its Gaussian-blurred
halo, and $\mathbf{c}_{w}$ and $\mathbf{c}_{p}$ denote the white and
purple color components, respectively.

% \paragraph{Ego-relative physical deployment.}
% \label{sec:physical_deployment}
% Digital search outputs the selected attack-position set $\mathcal{P}^*\in\{\{p_1^*\},\{p_1^*,p_2^*\}\}$ in the ego-centered BEV frame. To deploy physical emitters, we transform its $i$-th position $p_i^*=(x_i^{e},y_i^{e})$ using the ego pose $\mathbf{q}_{e}=(x_e,y_e,\theta_e)$:
% \begin{equation}
%     \begin{bmatrix}
%         x_i^{w}\\
%         y_i^{w}
%     \end{bmatrix}
%     =
%     \begin{bmatrix}
%         \cos\theta_e & -\sin\theta_e\\
%         \sin\theta_e & \cos\theta_e
%     \end{bmatrix}
%     \begin{bmatrix}
%         x_i^{e}\\
%         y_i^{e}
%     \end{bmatrix}
%     +
%     \begin{bmatrix}
%         x_e\\
%         y_e
%     \end{bmatrix},
%     \qquad i\in\{1,\ldots,|\mathcal{P}^*|\}.
%     \label{eq:ego_to_world}
% \end{equation}
% For each transformed position, we select the closest feasible deployment position $\tilde{p}_i$ from the set $\Omega_{\mathrm{phy}}$. We place one near-infrared emitter at each selected deployment position and activate them simultaneously. The attacked scene is captured by the same surround-view cameras and processed by the online map construction model used in digital evaluation. All physical evaluations are conducted in a closed test field under safety supervision and without public-road traffic.

\subsection{Ego-relative physical deployment}
\label{sec:physical_deployment}
The digital search returns one or two attack positions
$\mathcal{P}^*$ in the ego-centered BEV frame. Given the ego pose
$(x_e,y_e,\theta_e)$, each selected point $p_i^*$ is transformed to the
physical test field by
\begin{equation}
    p_i^{w}
    =
    R(\theta_e)p_i^*
    +
    \begin{bmatrix}
        x_e\\y_e
    \end{bmatrix},
    \label{eq:ego_to_world}
\end{equation}
where $R(\theta_e)$ is the 2D rotation matrix of the ego heading.

For each transformed point, we choose the nearest feasible deployment
position in $\Omega_{\mathrm{phy}}$, place a near-infrared emitter, and
activate all selected emitters simultaneously. The attacked scene is then
captured by the same surround-view cameras and processed by the online map
model used in digital evaluation. All tests are conducted in a closed
field under safety supervision.

%% file: sections/experiments.tex
\section{Experiments}
\begin{table*}[t]
\centering
\resizebox{\textwidth}{!}{
\begin{tabular}{l|cccc|cccc|cccc|cc}
\toprule
\multirow{2}{*}{\textbf{Model}} &
\multicolumn{4}{c|}{\textbf{Clean}} &
\multicolumn{4}{c|}{\textbf{Blind (ETA)}} &
\multicolumn{4}{c|}{\textbf{TwinIR (Ours)}} &
\multicolumn{2}{c}{\textbf{$\Delta$ mAP (pp)}} \\
\cmidrule(lr){2-5}\cmidrule(lr){6-9}\cmidrule(lr){10-13}\cmidrule(lr){14-15}
& Div. & Ped. & Bound. & mAP
& Div. & Ped. & Bound. & mAP
& Div. & Ped. & Bound. & mAP
& Blind-Clean & TwinIR-Clean \\
\midrule
MGMap     & 65.89 & 58.49 & 54.41 & 59.60 & 67.20 & 58.12 & 51.05 & 58.79 & 65.59 & 54.82 & 49.86 & 56.76 & -0.81 & -2.84 \\
MapTR     & 54.18 & 38.17 & 48.95 & 47.10 & 52.39 & 34.54 & 46.35 & 44.43 & 51.44 & 30.77 & 44.01 & 42.08 & -2.67 & -5.02 \\
DAMap     & 69.66 & 59.13 & 53.30 & 60.69 & 65.58 & 56.41 & 48.55 & 56.85 & 64.43 & 53.64 & 47.15 & 55.07 & -3.85 & -5.62 \\

\bottomrule
\end{tabular}
}
\caption{Map element AP(\%) under ETA attacks on asymmetric scenes.}
\label{tab:eta_main_full}
\end{table*}
\begin{table}[t]
\centering
\begin{tabular}{l|c|c|c}
\toprule
\textbf{Model} & \textbf{UPTR$_c$} & \textbf{Blind (ETA)} & \textbf{TwinIR (Ours)} \\
\midrule
MGMap     & 13.0 & 26.0 & 32.0 \\
MapTR     & 10.0 & 27.0 & 30.0 \\
DAMap     & 14.0 & 25.0 & 33.0 \\
\bottomrule
\end{tabular}
\caption{Planning impact of ETA on asymmetric scenes, measured by unsafe planned trajectory rate (UPTR, \%). UPTR$_c$ denotes the clean setting.}
\label{tab:eta_plan}
\end{table}
\subsection{Experimental Setup}
\label{sec:experimental_setup}
\paragraph{Dataset and models.}
We evaluate our attack on the nuScenes dataset
\cite{caesar2020nuscenes} using the same asymmetric-scene subset selected
by prior work \cite{lou2025asymmetry}. The subset contains 100 frames
chosen from 407 asymmetric scenes identified in the nuScenes validation
set. We evaluate three camera-based online HD map models: MapTR
\cite{liao2023maptr}, MGMap \cite{liu2024mgmap}, and DAMap
\cite{dong2025damap}. All models are tested on the same scenes and target
boundaries. Detailed information about the hardware, vehicle platform, and other results is provided in the technical supplement.

\paragraph{Attack settings.}
For each model, we compare the clean input, the original single-point
attack, and TwinIR, which selects up to two attack positions. All methods use
the same candidate regions, attack targets, and query budget.

\paragraph{Metrics.}For map
construction, we report
$\mathrm{AP}_{\mathrm{boundary}}$,
$\mathrm{AP}_{\mathrm{divider}}$,
$\mathrm{AP}_{\mathrm{ped}}$, and their mean mAP. For downstream planning,
we report Unreachable Goal Rate (UGR) for RSA and Unsafe Planned
Trajectory Rate (UPTR) for ETA. Higher UGR and UPTR indicate stronger
attack impact.
% \subsection{Main Results on RSA}
% \label{sec:rsa_results}
% Table~\ref{tab:rsa_main_full} reports the map construction results under
% RSA. TwinIR consistently causes a larger mAP reduction than the original
% single-point blinding attack on all three victim models. Specifically, the
% mAP drops by 8.18, 8.96, and 8.75
% percentage points on MGMap, MapTR, and DAMap, respectively, compared with
% 6.97, 6.99, and 6.51 points under the single-point attack. Relative to
% Blind, our method further reduces mAP by 1.21 points on MGMap, 1.97 points
% on MapTR, and 2.24 points on DAMap. This additional reduction is
% consistent with suppressing compensation-boundary cues that remain
% available after single-point perturbation.

% The largest degradation is observed on road-boundary prediction, which is
% the direct target of RSA. Compared with the clean setting, the boundary AP
% under TwinIR decreases by 12.02 points on MGMap, 11.44 points on
% MapTR, and 11.22 points on DAMap. Moreover, the attack also affects
% divider and pedestrian-crossing predictions, indicating that the induced
% boundary deformation propagates to other geometrically related map
% elements.

% Table~\ref{tab:rsa_plan} further shows the downstream planning impact.
% TwinIR increases UGR to 51\%, 55\%, and 53\% on MGMap, MapTR, and DAMap,
% respectively. Compared with Blind, this corresponds to an additional
% increase of 11, 11, and 12 percentage points. These results show that the
% stronger map deformation produced by TwinIR translates into a
% substantially higher probability of making valid planning goals
% unreachable.
\subsection{Main Results on RSA}
\label{sec:rsa_results}

Table~\ref{tab:rsa_main_full} reports the RSA results. TwinIR consistently
causes a larger mAP reduction than the single-point blinding attack across
all three models. The mAP drops by 8.18, 8.96, and 8.75 percentage points
on MGMap, MapTR, and DAMap, respectively, compared with 6.97, 6.99, and
6.51 points under Blind. This corresponds to an additional reduction of
1.21--2.24 points, consistent with suppressing compensation-boundary cues
that remain after single-point perturbation.

The strongest degradation appears in road-boundary prediction, whose AP
decreases by 12.02, 11.44, and 11.22 points from the clean setting on the
three models. TwinIR also affects divider and pedestrian-crossing
predictions, suggesting that the induced boundary deformation propagates
to related map elements.As shown in Table~\ref{tab:rsa_plan}, TwinIR increases UGR to 51\%, 55\%,
and 53\% on MGMap, MapTR, and DAMap, respectively, exceeding Blind by
11--12 percentage points. Thus, the stronger map deformation directly
increases the likelihood that valid planning goals become unreachable.

\subsection{Main Results on ETA}
\label{sec:eta_results}

% Table~\ref{tab:eta_main_full} reports the map construction performance
% under ETA. TwinIR consistently produces a larger mAP reduction than the
% original single-point blinding attack across all three victim models.
% Specifically, TwinIR decreases mAP by 2.84, 5.02, and 5.62 percentage
% points on MGMap, MapTR, and DAMap, respectively, compared with 0.81, 2.67,
% and 3.85 points under Blind. Relative to the single-point attack, our
% method further reduces mAP by 2.03 points on MGMap, 2.35 points on MapTR,
% and 1.78 points on DAMap.

% The degradation is particularly evident in road-boundary prediction,
% which directly determines the estimated turning geometry. Compared with
% the clean setting, TwinIR reduces boundary AP by 4.55 points on
% MGMap, 4.94 points on MapTR, and 6.15 points on DAMap. The consistent
% decrease across different architectures indicates that perturbing both the
% target boundary and the opposite-side straight compensation boundary is
% more effective than perturbing only the original turning region.

% Table~\ref{tab:eta_plan} shows that the stronger map deformation also
% increases downstream planning risk. TwinIR raises UPTR from the clean
% values of 13\%, 10\%, and 14\% to 32\%, 30\%, and 33\% on MGMap, MapTR,
% and DAMap, respectively. Compared with Blind, the corresponding increases
% are 6, 3, and 8 percentage points. Notably, ETA may cause only a moderate
% reduction in overall mAP while substantially increasing the unsafe
% trajectory rate, showing that global map accuracy alone does not fully
% capture the safety impact of targeted geometric attacks.
Table~\ref{tab:eta_main_full} reports the ETA results. TwinIR consistently
causes a larger mAP reduction than single-point blinding across all three
models. The mAP drops by 2.84, 5.02, and 5.62 percentage points on MGMap,
MapTR, and DAMap, respectively, compared with 0.81, 2.67, and 3.85 points
under Blind. This corresponds to an additional reduction of 1.78--2.35
points.

The strongest degradation appears in road-boundary prediction, whose AP
decreases by 4.55, 4.94, and 6.15 points from the clean setting. This
shows that perturbing both the target boundary and the straight
compensation boundary is more effective than attacking only the original
turning region. As shown in Table~\ref{tab:eta_plan}, TwinIR raises UPTR to 32\%, 30\%,
and 33\% on MGMap, MapTR, and DAMap, exceeding Blind by 3--8 percentage
points. These results also show that a moderate mAP reduction can still
produce a substantial increase in unsafe trajectories.
\subsection{Physical-World Infrared Visibility}
\label{sec:physical_ir_effect}

\begin{figure}[t]
  \centering
  \includegraphics[width=1\linewidth]{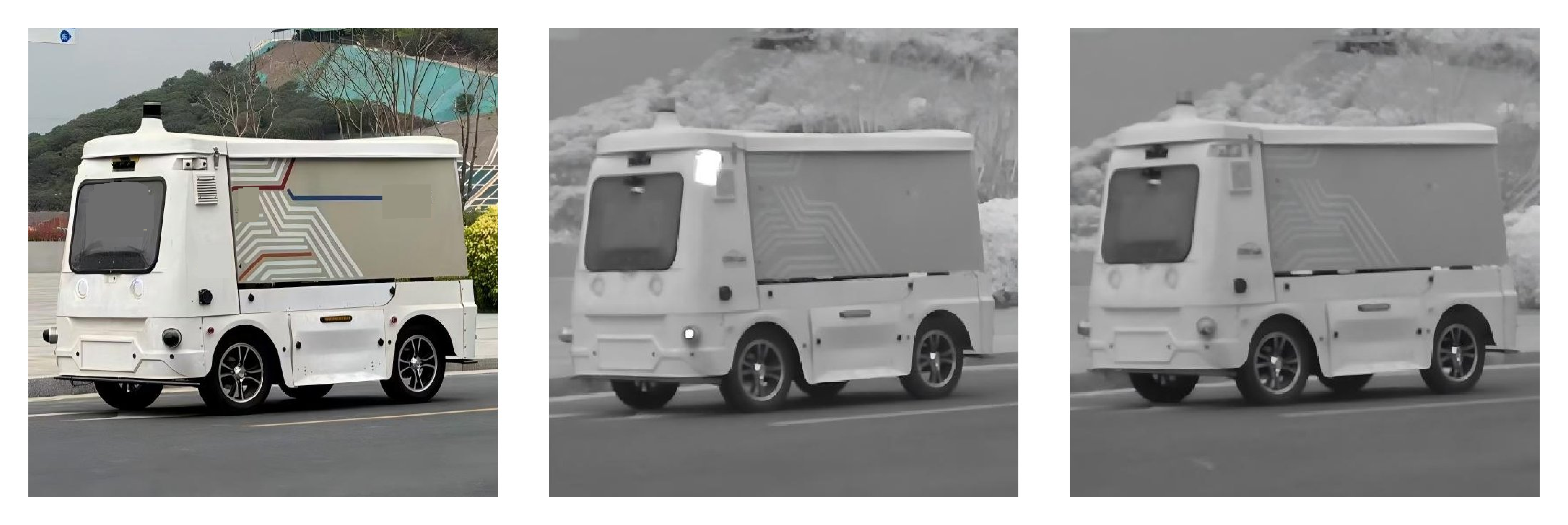}
\caption{
\textbf{Real-world visibility of infrared interference.}
(a) The attacked vehicle in a full-color image shows little visible change.
(b) The infrared-sensitive camera clearly captures the attack response.
(c) The unattacked infrared image serves as the clean baseline.
}
    \label{fig:physical_ir_effect}
\end{figure}

We first examine whether the infrared attack remains inconspicuous in
visible images while producing a clear camera response.
Figure~\ref{fig:physical_ir_effect} compares the same vehicle under three
imaging conditions.

As shown in Fig.~\ref{fig:physical_ir_effect}(a), activating the infrared
source causes little visible change in the full-color image. In contrast,
the infrared-sensitive camera clearly captures the injected response in
Fig.~\ref{fig:physical_ir_effect}(b). Figure~\ref{fig:physical_ir_effect}(c)
shows the clean infrared baseline without attack activation. Comparing
(b) with (c) confirms that the additional artifact is introduced by the
infrared source, while comparing (a) with (b) demonstrates its limited
visible-domain appearance.

\begin{figure}[t]
\centering
\includegraphics[trim={0cm 6cm 2.3cm 0cm}, clip, width=\linewidth]{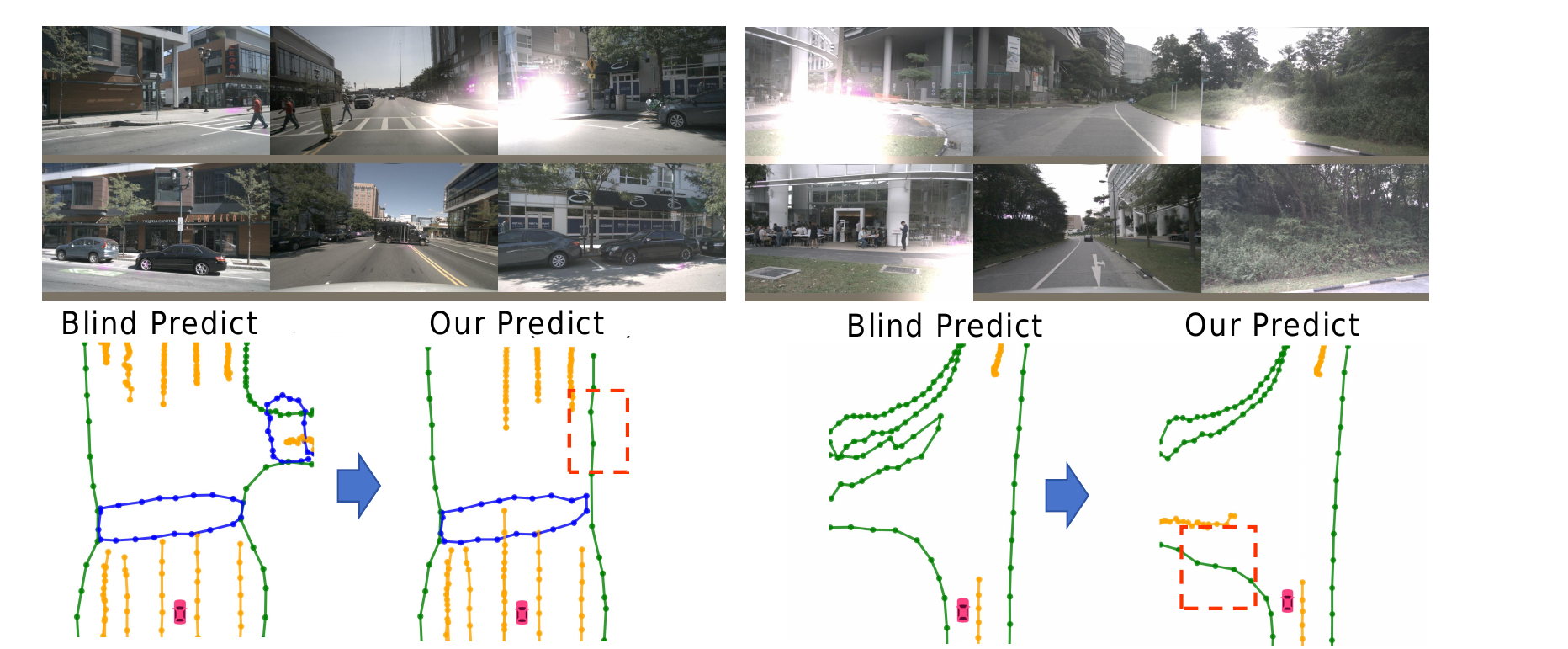}
\caption{
\textbf{Visualization results of TwinIR in nuScenes.}
Compared with single-point blinding, our dual-point attack successfully
achieves the target deformation for both RSA (left) and ETA (right).
}
\label{fig:qualitative_results}
\end{figure}
\subsection{Visualization Results}
\label{sec:qualitative_results}

Figure~\ref{fig:qualitative_results} presents representative visualization
results of our dual-point attack. The left example corresponds to RSA,
where the original single-point blinding attack fails to fully remove the
turning structure, while our dual-point attack successfully suppresses the
complementary boundary evidence and induces the desired straightening
effect. The right example corresponds to ETA, where Blind cannot move the
turn sufficiently early, but our method further perturbs the reference
boundary and successfully shifts the predicted turn toward the ego vehicle.
These examples directly mirror the failure cases in the motivation
section and show that the secondary attack position effectively improves attack
coverage by removing alternative geometric cues used by the model.
\begin{figure}[t]
\centering
\includegraphics[width=\linewidth, trim=0cm 1cm 14cm 0cm, clip]{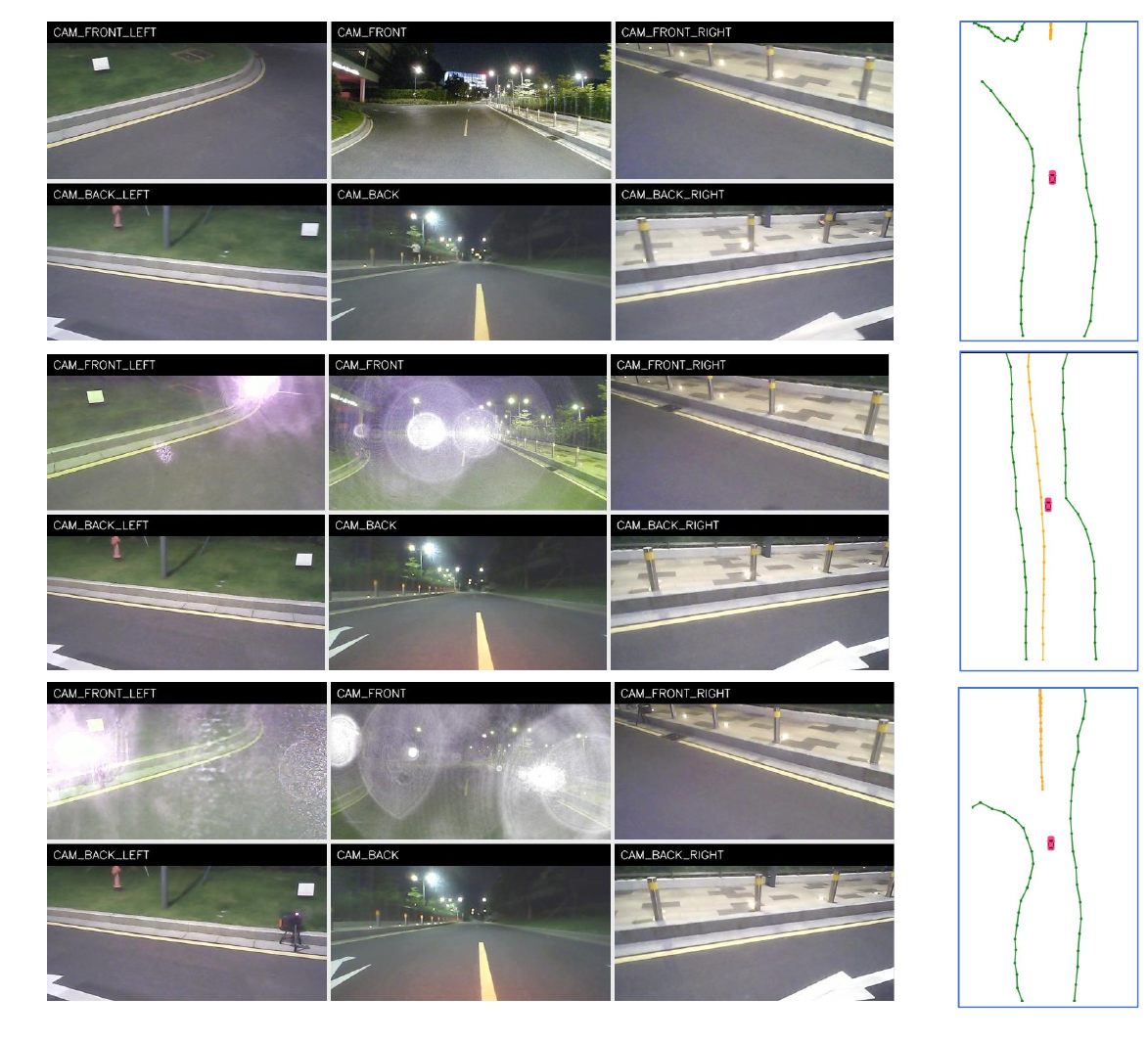}
\caption{
\textbf{Controlled physical-world results of TwinIR.}
From top to bottom: clean input, RSA, and ETA. The six surround-view camera
images are shown on the left, and the corresponding online map predictions
are shown on the right.
}
\vspace{-5px}
\label{fig:physical_attack}
\end{figure}
\subsection{Controlled Physical-World Evaluation}
\label{sec:physical_results}

We further evaluate TwinIR on a real vehicle in a controlled test field.
Figure~\ref{fig:physical_attack} presents the six surround-view camera
images and the corresponding online map predictions under the clean, RSA,
and ETA settings. The clean scene produces the original road geometry.
Under RSA, the infrared interference alters the target boundary and makes
the predicted road structure straighter. Under ETA, the predicted turning
location is shifted toward the ego vehicle. The infrared responses are
clearly captured by the onboard cameras at the ego-relative positions
selected during digital optimization. The resulting deformations are also
consistent with the corresponding digital attack objectives. Moreover, the
attack remains effective after transferring the optimized positions from
the digital scene to the physical test field. This indicates that the
selected point pairs are not limited to simulation-specific image patterns.
These results demonstrate that the digitally selected dual-point positions
can be transferred to the physical environment and induce the intended map deformations.

%% file: sections/related_work.tex
\section{Related Work}

\subsection{Online Vectorized HD Map Construction}

Online HD map construction predicts local road maps directly from onboard sensor observations. Early approaches, such as HDMapNet, first generate bird's-eye-view semantic maps and then convert them into vectorized elements through post-processing \cite{li2022hdmapnet}. More recent methods perform end-to-end vectorized prediction. VectorMapNet autoregressively generates map-element polylines \cite{liu2023vectormapnet}, while MapTR and MapTRv2 represent each element as a structured point set and jointly learn instance- and point-level correspondences \cite{liao2023maptr,liao2025maptrv2}. Other approaches introduce stronger geometric and structural priors, including B'ezier-curve representations \cite{qiao2023bemapnet}, pivotal-point modeling \cite{ding2023pivotnet}, geometry-aware relation learning \cite{zhang2024gemap}, and lane-segment representations that combine paired boundaries, centerlines, and topology \cite{li2024lanesegnet}.

Temporal methods further exploit historical observations to improve mapping consistency. StreamMapNet propagates map queries across frames \cite{yuan2024streammapnet}, while MapTracker maintains temporal memories of map elements \cite{chen2024maptracker}. Although these methods improve robustness to occlusion and incomplete observations, their reliance on paired boundaries, geometric relations, and temporal context also enables a corrupted map element to be reconstructed from residual geometric cues. Our work investigates how such structural dependencies affect the effectiveness of physical attacks.

\subsection{Adversarial Attacks on Autonomous Driving}

Physical-world adversarial attacks have been extensively studied for autonomous-driving perception. Adversarial stickers can manipulate traffic-sign recognition \cite{eykholt2018robust}, road-surface perturbations can mislead lane-detection and lane-centering systems \cite{jing2021toogood,sato2021dirty}, and adversarial objects or spoofed signals can compromise LiDAR-based detectors \cite{tu2020physically,sun2020robustlidar}. Multi-sensor fusion systems are also vulnerable when attacks preserve consistency across camera and LiDAR observations \cite{cao2021invisible,hallyburton2022fusion}.

\subsection{Attacks on Online HD Map Construction}

Lou et al. identify a symmetry bias in online HD map construction: when critical geometric cues in an asymmetric scene are disrupted, the prediction may follow cues from an opposite or reference boundary \cite{lou2025asymmetry}. They exploit this bias with camera blinding and adversarial patches placed at a single optimized position. Wang et al. propose MIRAGE, which uses conditional diffusion to discover plausible semantic variations, such as shadows and wet roads, that preserve road topology while inducing boundary removal or injection \cite{wang2026mirage}.

These studies expose complementary threats, but they do not examine coordinated multi-point physical illumination. Building on the observation that one boundary can influence another, we analyze unsuccessful single-point attacks and find that the same cross-boundary dependence can preserve the correct geometry: an unaffected compensation boundary retains geometric cues for the original turn and compensates for the perturbed region. TwinIR therefore perturbs the target boundary and conditionally suppresses the compensation boundary with objective-conditioned two-point near-infrared illumination while preserving the guiding boundary.

%% file: sections/hardware_appendix.tex
% Include this file from the main paper with:
% \appendix
% \input{sections/TwinIR_hardware_appendix_final}

\section{Hardware for Controlled Physical Experiments}
\label{app:hardware}

This section describes the hardware used in the controlled physical
experiments. The setup consisted of a test vehicle equipped with six
surround-view cameras, auxiliary obstacle-detection sensors, and one or two
near-infrared laser emitters powered independently of the vehicle. All tests
were conducted in a restricted test field without public-road traffic or
uninvolved road users.

\subsection{Test Vehicle and Camera System}

The test vehicle comprised a TongLing W102 drive-by-wire chassis and an
in-house-fabricated upper structure. The vehicle platform was capable of
autonomous driving, although its speed did not exceed
$10\,\mathrm{km\,h^{-1}}$ during the experiments. The platform used six
LRCP10620 USB cameras, supplied by a third-party vendor and equipped with
Sony $1/2.7$-inch CMOS sensors, to collect the surround-view images evaluated
by the online HD map construction system. The front and lateral cameras
targeted in the physical experiments were mounted approximately $1.5$~m
above the ground.

Each camera recorded $640\times480$ 8-bit images at 10 frames per second.
The pixel size was
$3.0\,\mu\mathrm{m}\times3.0\,\mu\mathrm{m}$, the signal-to-noise ratio was
39~dB, and the dynamic range was 69~dB. The nominal fields of view were
$75^\circ$ for the front and lateral cameras and $100^\circ$ for the rear
cameras. All cameras used rolling shutters with automatic exposure and gain
control. No IR-cut filter was installed; therefore, the observed
near-infrared response arose from the native spectral sensitivity of the
CMOS sensors.

\paragraph{Onboard computing and data logging.}
The vehicle carried an industrial computer with an Intel Core i9-14900 CPU,
an NVIDIA GeForce RTX 4060 Ti GPU, 32~GB of memory, and a 1-TB SATA
solid-state drive. A separate 2-TB NVMe solid-state drive connected over
PCIe 4.0 was used for sensor data collection and logging. During the
experiments reported in this work, the onboard computer served only as a
data-acquisition platform and was not used for model training, online model
inference, or attack optimization.

\paragraph{Auxiliary safety sensors.}
The vehicle was additionally equipped with one 32-channel RoboSense
Helios 32 LiDAR, three 32-channel RoboSense RS-Bpearl blind-spot LiDARs,
and 12 ultrasonic sensors. These sensors provided an auxiliary
obstacle-detection safety layer. When the safety controller detected an
obstacle in the vehicle's direction of travel, it automatically issued an
emergency-stop command to the drive-by-wire chassis to prevent a collision
during the experiments. These auxiliary sensors were used solely for
operational safety and were not used as inputs to the camera-based online HD
map construction models evaluated in this work.

\begin{figure}[htbp]
\centering % 居中
\vspace{-10pt} % 吃掉上方空白
\includegraphics[width=0.48\textwidth]{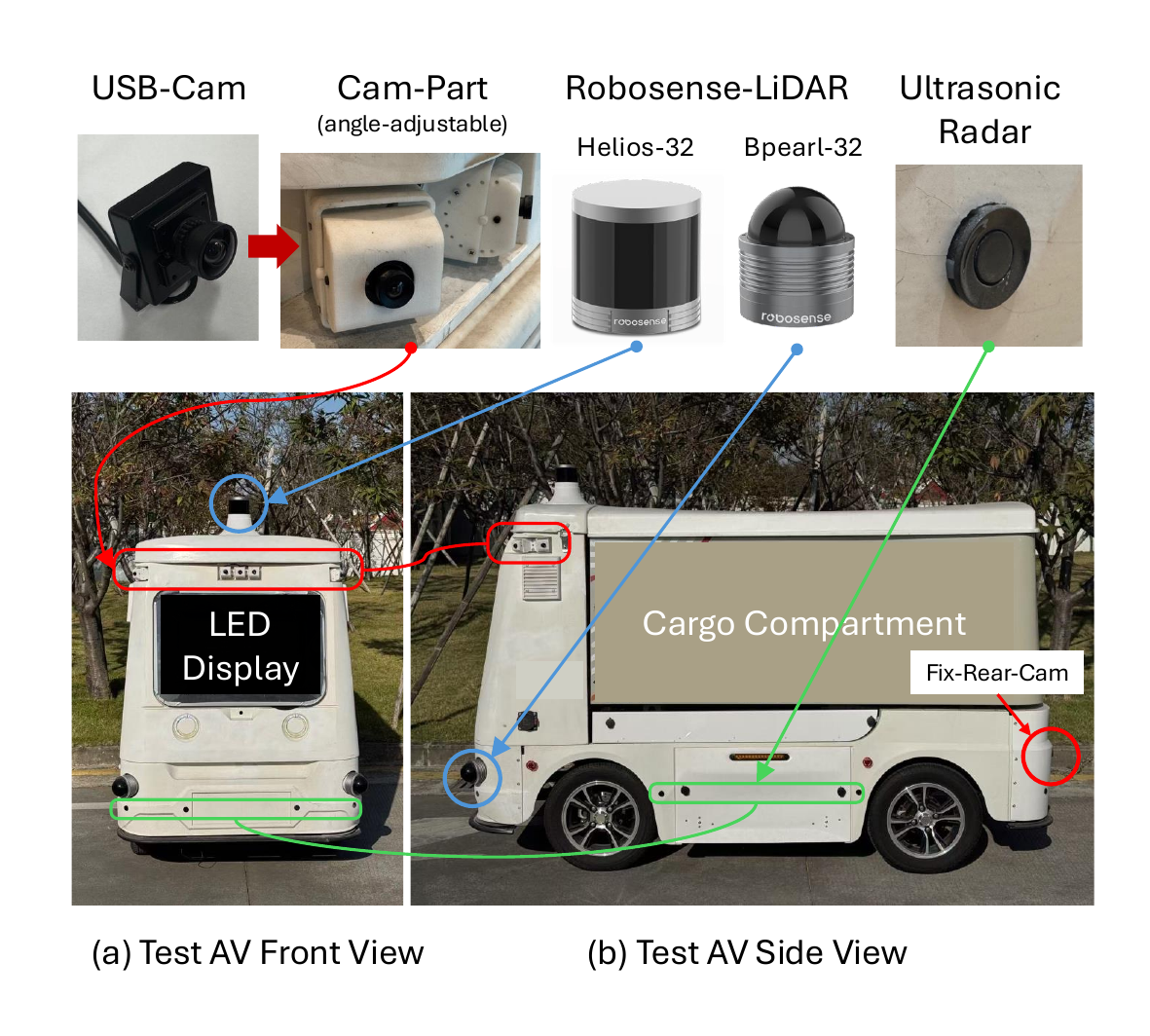}
\vspace{-24pt}  % 图片到标题之间收紧
\caption{Tested AV}
\vspace{-16pt}  % 标题下方空白缩小
\label{fig:appendix-vehicle&sensor-image} % 标签，正文可用 \ref{fig:arch} 引用图号
\end{figure}

\subsection{Near-Infrared Illumination System}

Each physical attack position was implemented using one OXlasers near-infrared
laser-diode emitter. Each emitter had a center wavelength of 808~nm and a
manufacturer-rated optical power of 500~mW; the emitted optical power was
not independently measured. Each emitter was driven through a 0--5-V
pulse-width-modulation (PWM) interface supporting frequencies below
18~kHz. During the attacked recordings, the duty cycle was set to 100\%,
yielding effectively continuous output at the maximum configured level.
Therefore, phase synchronization with the cameras was unnecessary. The spot
diameter was approximately 30~cm at a
source-to-target distance of 10~m.

Each emitter was mounted on a hydraulic tripod at approximately 0.35~m
above the ground, and the source-to-target distance ranged from 3 to 10~m.
One emitter was used for a retained one-point configuration, whereas two
identical emitters were activated simultaneously for a retained two-point
TwinIR configuration.

The emitters were powered independently of the test vehicle using a portable
power station. Each laser unit incorporated separate power-on and
operating-status indicators. The wiring harness remained physically
disconnected until the system was ready for controlled activation.

\paragraph{Infrared alignment and observation camera.}
During the controlled physical attacks, a separate USB 2.0 infrared camera
was used to assist in aiming the laser emitters at the target regions and
observing the resulting infrared response. The camera was fitted with an
800--950-nm band-pass filter to limit the detected spectrum primarily to
the near-infrared range. This camera was used only as physical-attack
instrumentation; it was not involved in model training or attack-position
optimization and was not an input to the victim online HD map construction
models. The visible-light image in Fig.~5(a) was captured using a consumer
smartphone camera, whereas the attacked and clean infrared images in
Figs.~5(b) and 5(c), respectively, were captured using this infrared camera.

\begin{figure}[htbp]
\centering % 居中
\vspace{-10pt} % 吃掉上方空白
\includegraphics[width=0.48\textwidth]{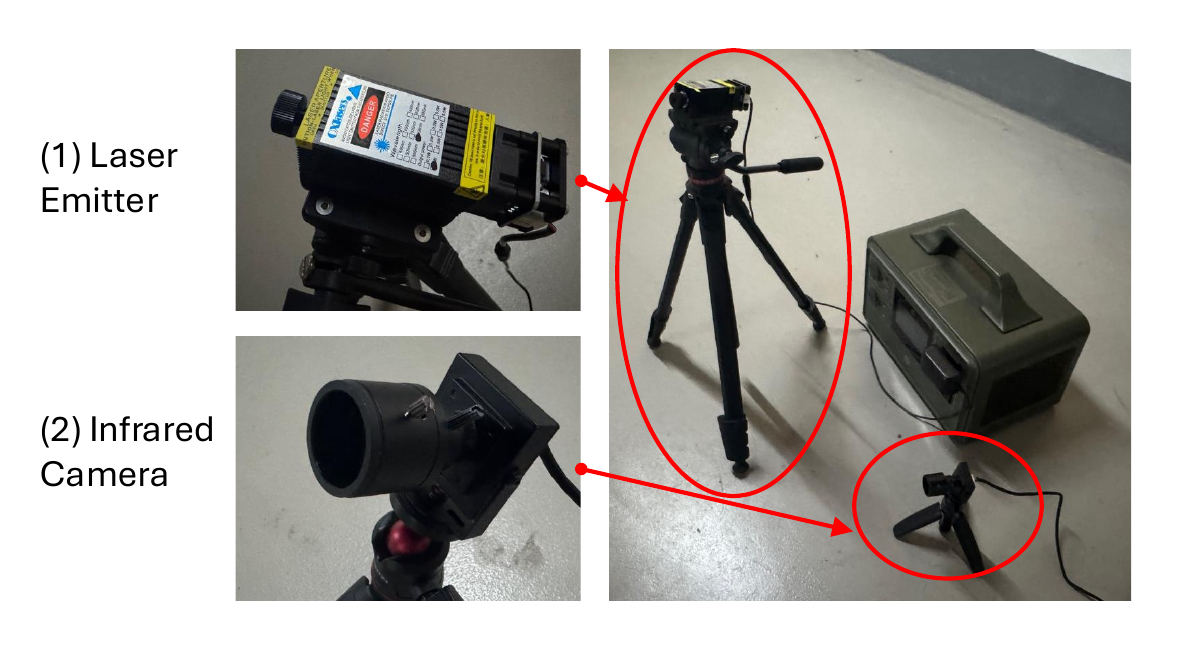}
\vspace{-24pt}  % 图片到标题之间收紧
\caption{Attacking Component}
\vspace{-16pt}  % 标题下方空白缩小
\label{fig:appendix-vehicle&sensor-image} % 标签，正文可用 \ref{fig:arch} 引用图号
\end{figure}

\subsection{Deployment and Safety Protocol}

For each scene, we fixed the vehicle pose and transformed the retained
ego-frame attack positions into the test-field coordinate system using the
ego-to-world coordinate transformation described in the main paper. We then
placed the emitters at the closest feasible deployment positions and oriented them
toward the designated regions of the vehicle. Clean and attacked recordings
used the same vehicle pose and camera configuration, with automatic exposure
and gain control enabled under both conditions.

For experimental safety management, the emitters were treated as Class 3B
laser devices; this classification was not independently verified. The
emitters were operated only within a restricted test area. Before power was
applied, all participating personnel withdrew from the designated hazardous
area and confirmed that it was clear. Consequently, no participating
personnel were present in the beam path while the laser emitters were
energized. After each recording, the emitters were de-energized before
personnel re-entered the restricted area.

\begin{figure}[htbp]
\centering % 居中
\vspace{-10pt} % 吃掉上方空白
\includegraphics[width=0.48\textwidth, trim=0cm 6cm 10cm 0cm, clip]{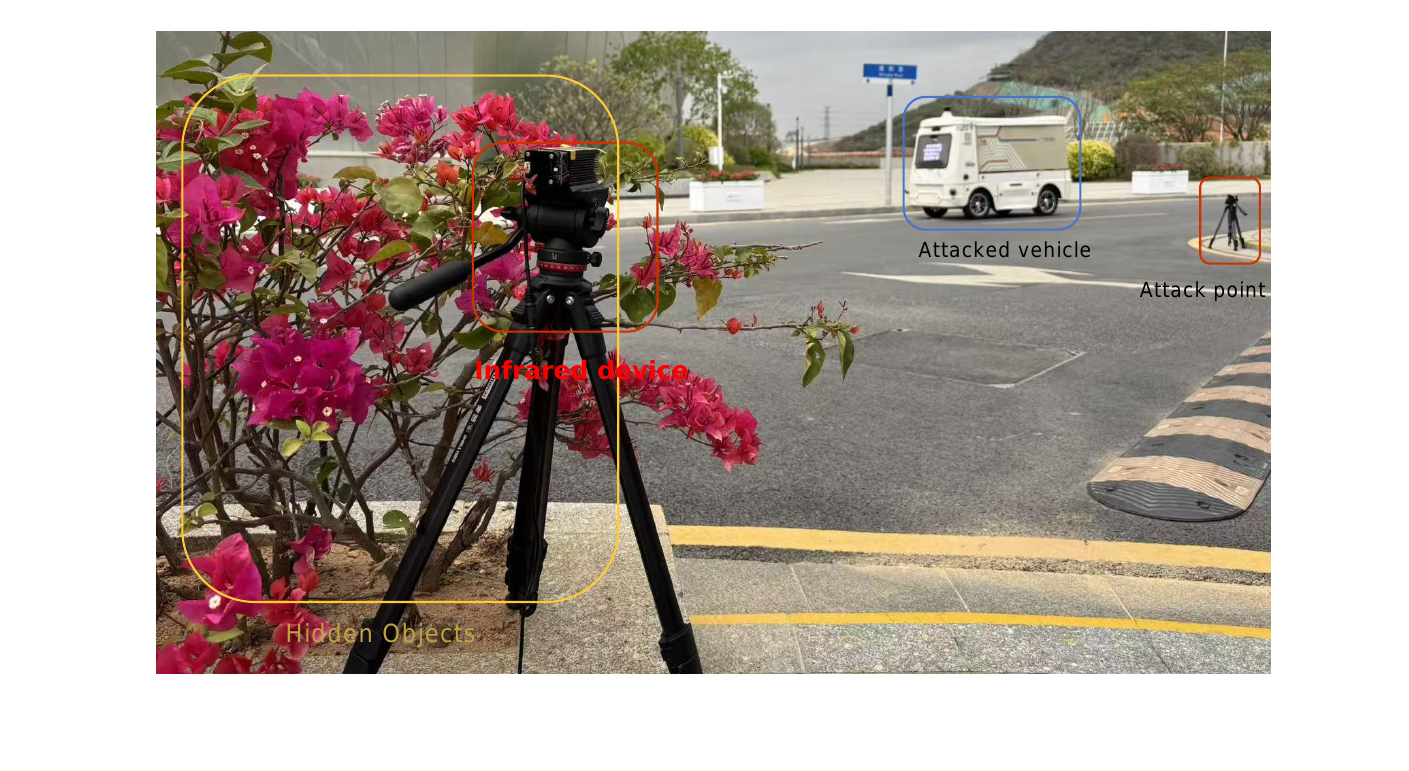}
\vspace{-24pt}  % 图片到标题之间收紧
\caption{Physical deployment of TwinIR, where the infrared device is concealed by roadside objects while maintaining a clear line of sight to the attacked vehicle.}
\vspace{-16pt}  % 标题下方空白缩小
\label{fig:physical_setup} % 标签，正文可用 \ref{fig:arch} 引用图号
\end{figure}
\subsection{Stealthiness of Physical Deployment}
\label{sec:deployment_stealthiness}

Figure~\ref{fig:physical_setup} illustrates the physical deployment of
TwinIR. To reduce visual exposure, the infrared device is placed behind
existing roadside objects and vegetation, while only the emitter is
directed toward the attacked vehicle. The attack point is deployed using
a small roadside tripod at the optimized position, requiring no physical
contact with or modification to the vehicle. From the vehicle and normal
roadside viewing directions, most supporting hardware is therefore
occluded by the surrounding environment. This setup demonstrates that
TwinIR can be integrated into an ordinary roadside scene with limited
visible infrastructure, while still maintaining a clear line of sight
between the infrared emitter and the target cameras.

\begin{table}[t]
\centering
\resizebox{0.5\textwidth}{!}{
\begin{tabular}{l|c|cccc}
\toprule
\textbf{Model} & \textbf{\#Points $k$} & \textbf{Divider} & \textbf{Ped.} & \textbf{Bound.} & \textbf{mAP} \\
\midrule
\multirow{5}{*}{MGMap}
& 1 & 57.75 & 53.74 & 46.40 & 52.63 \\
& 2 & 56.33 & 55.53 & 42.39 & 51.41 \\
& 3 & 55.91 & 55.96 & 41.85 & 51.24 \\
& 4 & 56.29 & 55.19 & 41.41 & 50.96 \\
& 5 & 56.68 & 54.41 & 41.41 & 50.83 \\
\midrule
\multirow{5}{*}{DAMap}
& 1 & 61.64 & 55.58 & 45.34 & 54.18 \\
& 2 & 60.14 & 53.61 & 42.08 & 51.94 \\
& 3 & 60.61 & 52.56 & 42.06 & 51.74 \\
& 4 & 61.25 & 51.84 & 41.73 & 51.61 \\
& 5 & 60.19 & 51.38 & 42.14 & 51.24 \\
\midrule
\multirow{5}{*}{MapTR}
& 1 & 44.12 & 36.06 & 40.16 & 40.11 \\
& 2 & 42.77 & 34.15 & 37.51 & 38.14 \\
& 3 & 42.38 & 33.49 & 38.04 & 37.97 \\
& 4 & 42.46 & 33.01 & 37.82 & 37.77 \\
& 5 & 41.69 & 33.21 & 37.71 & 37.54 \\
\bottomrule
\end{tabular}
}
\caption{Multi-point RSA attack. $k$ denotes the number of light points. Values are AP(\%) and mAP(\%).}
\label{tab:multipoint_rsa}
\end{table}
\subsection{Effect of the Number of Attack Points}
\label{sec:multipoint_ablation}

Tables~\ref{tab:multipoint_rsa} and~\ref{tab:multipoint_eta} study the
effect of increasing the number of infrared attack points from one to
five. Across both objectives and all victim models, the largest marginal
degradation is consistently obtained when increasing the number of points
from one to two. Under RSA, adding the second point further reduces mAP by
1.22, 2.24, and 1.97 percentage points on MGMap, DAMap, and MapTR,
respectively. In contrast, increasing the number of points from two to
five provides only an additional reduction of 0.58--0.70 points.

A similar trend is observed under ETA. The second attack point decreases
mAP by 3.62 points on MGMap, 2.08 points on DAMap, and 1.42 points on
MapTR. However, adding three more points after $k=2$ reduces mAP by only
0.69--0.71 additional points. Therefore, although three to five points
occasionally achieve slightly lower absolute mAP, their marginal benefits
are substantially smaller than that of the second point.

These results indicate that the attack effect becomes largely saturated
after two points. The first point perturbs the target boundary, while the
second suppresses the complementary geometric cues provided by the
compensation boundary. Once these two major sources of road-geometry
evidence are jointly disrupted, additional points mainly perturb
overlapping local features and provide limited further benefit. The
consistent saturation trend across different model architectures and
attack objectives also suggests that this behavior is not specific to a
single victim model.

Considering that additional emitters increase deployment, calibration,
and concealment costs, $k=2$ provides the best trade-off between attack
effectiveness and physical practicality. This observation supports the
at-most-two-point design of TwinIR.
\begin{table}[t]
\centering
\resizebox{0.5\textwidth}{!}{
\begin{tabular}{l|c|cccc}
\toprule
\textbf{Model} & \textbf{\#Points $k$} & \textbf{Divider} & \textbf{Ped.} & \textbf{Bound.} & \textbf{mAP} \\
\midrule
\multirow{5}{*}{MGMap}
& 1 & 64.88 & 56.07 & 50.84 & 57.27 \\
& 2 & 59.21 & 53.78 & 47.96 & 53.65 \\
& 3 & 58.92 & 53.83 & 47.06 & 53.27 \\
& 4 & 58.10 & 54.77 & 46.61 & 53.16 \\
& 5 & 57.80 & 54.57 & 46.48 & 52.95 \\
\midrule
\multirow{5}{*}{DAMap}
& 1 & 64.80 & 55.49 & 47.83 & 56.04 \\
& 2 & 63.68 & 51.80 & 46.39 & 53.96 \\
& 3 & 64.11 & 51.86 & 45.50 & 53.82 \\
& 4 & 62.89 & 51.67 & 45.82 & 53.46 \\
& 5 & 63.42 & 51.24 & 45.09 & 53.25 \\
\midrule
\multirow{5}{*}{MapTR}
& 1 & 51.86 & 35.34 & 46.08 & 44.43 \\
& 2 & 50.07 & 33.38 & 45.57 & 43.01 \\
& 3 & 49.22 & 33.71 & 45.77 & 42.90 \\
& 4 & 48.30 & 34.11 & 45.36 & 42.59 \\
& 5 & 48.84 & 33.25 & 44.87 & 42.32 \\
\bottomrule
\end{tabular}
}
\caption{Multi-point ETA invisible-dual attack. $k$ denotes the number of light points. Values are AP(\%) and mAP(\%).}
\label{tab:multipoint_eta}
\end{table}

\subsection{Implementation Details}

\paragraph{Hardware.}
All experiments are conducted on a workstation equipped with a single NVIDIA RTX~A4000 GPU (16\,GB), an Intel Core~i7-12700 CPU (20 logical cores), and 62\,GB of system memory.
Inference and attack optimization are executed under CUDA with PyTorch in the corresponding model environments (MapTR/DAMap share one environment; MGMap uses a separate environment).
Unless otherwise specified, each attack run uses one GPU with batch size~1.

\paragraph{Victim models.}
We evaluate three representative camera-based online HD map constructors with publicly released checkpoints and keep their official architectures, BEV settings, and pretrained weights unchanged:
\begin{itemize}
    \item \textbf{MapTR}~\cite{liao2023maptr}: tiny variant with ResNet-50 backbone, BEV pooling, trained for 24 epochs.
    The perception range is $[-15,15]\times[-30,30]$\,m on the ground plane, with BEV resolution $H{\times}W{=}200{\times}100$, $N{=}50$ map queries, and a single-frame temporal queue ($q{=}1$).
    Clean mAP on our asymmetric NuScenes subset is 47.10\%.
    \item \textbf{DAMap}~\cite{dong2025damap}: MapTRv2-based tiny model with ResNet-50 backbone, trained for 24 epochs.
    Clean mAP is 60.69\%.
    \item \textbf{MGMap}~\cite{liu2024mgmap}: ResNet-50 camera-only model trained for 30 epochs.
    Clean mAP is 59.60\%.
\end{itemize}
For all models, map elements are decoded into divider, pedestrian-crossing, and boundary polylines and scored by Chamfer-distance mAP at thresholds $\{0.5,1.0,1.5\}$\,m.

\paragraph{Attack rendering and search hyperparameters.}
Physical light effects are rendered as additive purple--white lens flares in image space.
Unless stated otherwise, the flare power is $P{=}3000$, the emitter height is fixed at $z{=}-1.84$\,m, and candidate visibility is scored with a maximum beam angle of $40^\circ$.
For Blind (single-point) search, we sample up to $400$ candidate locations at interval $0.5$\,m, with $4$ discrete heights, local resampling ($2$ samples per location within a $1.0$\,m range), and lateral lane offset $0.3$\,m.
For Invisible Dual, Stage~1 reuses the strongest Blind location as $L_1$; Stage~2 evaluates $N_2{=}400$ candidates for $L_2$.
RSA places $L_2$ on the edge touched by the straightened target and uses up to two adaptive expansion rounds (initial radius $22$\,m, expansion factor $1.5$), with a minimum inter-light distance of $3.0$\,m.
ETA samples $L_2$ over the complete reference edge without expansion and does not enforce the same minimum spacing.
A dual-point configuration is accepted only when its attack loss strictly improves over the Blind baseline; otherwise the Blind solution is retained.
Multi-point ablations vary the number of lights $k\in\{1,2,3,4,5\}$ under the same rendering and evaluation protocol, with $k{=}2$ as the default operating point.
Downstream planning uses Hybrid~A$^*$ with a collision threshold of $0.5$\,m; RSA is evaluated by UGR and ETA by UPTR.